\documentclass{article}
\usepackage{ijcai22}

\usepackage{times}
\usepackage{soul}
\usepackage{url}
\usepackage[hidelinks]{hyperref}
\usepackage[utf8]{inputenc}
\usepackage[small]{caption}
\usepackage{graphicx}
\usepackage{amsmath}
\usepackage{amsthm}
\usepackage{booktabs}
\usepackage{algorithm}
\usepackage{algorithmic}
\usepackage{multirow}
\usepackage{diagbox}
\usepackage{stfloats}
\usepackage{fancyhdr}
\title{Dite-HRNet: Dynamic Lightweight High-Resolution Network for Human Pose Estimation}

\author{
Qun Li$^1$\and
Ziyi Zhang$^1$\and
Fu Xiao$^1$\footnote{Contact Author}\and
Feng Zhang$^1$\And
Bir Bhanu$^2$\\ 
\affiliations
$^1$School of Computer Science, Nanjing University of Posts and Telecommunications, Nanjing, China\\ 
$^2$Department of Electrical and Computer Engineering, University of California at Riverside, CA, USA\\
\emails
\{liqun, xiaof, zhangfeng01\}@njupt.edu.cn,
zhangziyi\_njupt@hotmail.com,
bhanu@ee.ucr.edu
}

\begin{document}

\maketitle

\begin{abstract}
A high-resolution network exhibits remarkable capability in extracting multi-scale features for human pose estimation, but fails to capture long-range interactions between joints and has high computational complexity. To address these problems, we present a Dynamic lightweight High-Resolution Network (Dite-HRNet), which can efficiently extract multi-scale contextual information and model long-range spatial dependency for human pose estimation. Specifically, we propose two methods, dynamic split convolution and adaptive context modeling, and embed them into two novel lightweight blocks, which are named dynamic multi-scale context block and dynamic global context block. These two blocks, as the basic component units of our Dite-HRNet, are specially designed for the high-resolution networks to make full use of the parallel multi-resolution architecture. Experimental results show that the proposed network achieves superior performance on both COCO and MPII human pose estimation datasets, surpassing the state-of-the-art lightweight networks. Code is available at: \url{https://github.com/ZiyiZhang27/Dite-HRNet}.
\end{abstract}

\section{Introduction}
Human pose estimation is a task that requires both accuracy and efficiency, especially when it encounters real-time applications on devices with limited resources. In human pose estimation, high-resolution representation is necessary for accuracy, which brings difficulties to achieve high efficiency. 

While High-Resolution Network (HRNet) \cite{sun:hrnet} achieves remarkable performance in human pose estimation, it has high computational complexity. To make it lightweight, Small HRNet \cite{wang:smallhrnet} reduces the width and depth of the network, but such reduction introduces performance degradation. Lightweight High-Resolution Network (Lite-HRNet) \cite{yu:litehrnet} shows that lightweight high-resolution network can achieve satisfactory performance by replacing each residual block in Small HRNet with an efficient CNN block. However, it remains unclear whether such static block that is independent of the inputs can provide the optimal trade-off between the network performance and complexity for high-resolution networks, as some operations may have different effects at different locations of the network and with inputs of different sizes \cite{cui:pplcnet}. The problem is that input-dependent component units may perform more efficiently in high-resolution networks than the input-independent counterparts. Our work focuses on designing dynamic blocks to develop a lightweight high-resolution network with dynamic representations.

Recent research \cite{wang:nonlocal} has demonstrated the benefits of long-range spatial dependency in human pose estimation. The general concept of long-range spatial dependency is the global understanding of spatial information in a large field of view. Previous high-resolution networks \cite{sun:hrnet,cheng:higherhrnet,wang:smallhrnet,yu:litehrnet} mainly rely on the deeply stacked convolution layers in parallel branches that extract multi-scale features to build spatial dependency, but the capacity of lightweight networks can be severely affected by the restricted widths and depths. Hence, another problem is how to enhance lightweight high-resolution networks with more efficient methods for modeling long-range spatial dependency.

To address the above problems, we present a Dynamic lightweight High-Resolution Network (Dite-HRNet). We adopt the architecture similar to HRNet \cite{sun:hrnet} and develop two dynamic lightweight blocks to improve the overall efficiency. First, we propose a Dynamic Split Convolution (DSC), which can dynamically extract multi-scale contextual information and optimize the trade-off between its capacity and complexity by two hyper-parameters, and thus is more effective and flexible than the standard convolution. Then, we introduce long-range spatial dependency into the high-resolution network by designing a novel Adaptive Context Modeling (ACM) method that enables the model to learn both local and global patterns of human poses. Finally, we embed DSC and ACM into two dynamic lightweight blocks, which are specially designed for the high-resolution network to make full use of the parallel multi-resolution architecture, as the basic component units of our Dite-HRNet.

The main contributions of this paper can be summarized as follows: (1) We propose a novel lightweight network named Dite-HRNet that can efficiently extract multi-scale contextual information and model long-range spatial dependency for human pose estimation. (2) We design a DSC method to build multi-scale contextual relationship and an ACM method to further strengthen long-range spatial dependency. In addition, we accomplish a joint embedding of DSC and ACM into two dynamic lightweight blocks, which are designed as the basic components of Dite-HRNet. (3) In our experiments, Dite-HRNet achieves superior trade-off between the network performance and complexity on both COCO and MPII human pose estimation datasets.

\section{Related Work}

\paragraph{Lightweight Human Pose Estimation.} Recent studies \cite{wang:smallhrnet,yu:litehrnet} focus on improving the efficiency of huamn pose estimation networks. Small HRNet \cite{wang:smallhrnet} reduces the width and depth of the original large version of HRNet \cite{sun:hrnet}. Lite-HRNet \cite{yu:litehrnet} exploits the potential of lightweight high-resolution network by replacing each residual block in the Small HRNet with a conditional channel weighting block, which adopts channel attention mechanism \cite{wang:senet} by replacing high-cost $1 \times 1$ convolutions in shuffle block \cite{ma:shuffle2} with channel weighting operations.

\paragraph{Efficient CNN Blocks.} Efficient CNN blocks have been widely used in many efficient CNN architectures \cite{howard:mobile1,howard:mobile3,zhang:shuffle1}, which aim to maximize the capacity of models under limited computational cost. MobileNetV2 \cite{sandler:mobile2} introduces the inverted residual block, which improves both accuracy and efficiency over traditional bottleneck blocks. Shuffle block in ShuffleNetV2 \cite{ma:shuffle2} performs convolutions only on half of the channels with a competitive performance, due to the channel split operation that splits the features by channels and the channel shuffle operation that enhances the information exchange across channels. MixNet \cite{tan:mixnet} combines multiple convolution layer into a mixed depth-wise convolution, which is a drop-in replacement of vanilla depth-wise convolution in CNN blocks. Sandglass block \cite{zhou:next} flips the structure of the inverted residual block, reducing the information loss without any additional computational cost.

\paragraph{Spatial Dependency Modeling.} Long-range spatial dependency can be spontaneously captured in the repeated convolution layers, which have better performance with large convolution kernels. However, deeply stacked convolution layers and large convolution kernels are both costly and inflexible to be integrated in lightweight networks. Non-local network \cite{wang:nonlocal} adopts a self-attention mechanism to model pixel-wise spatial relations in a single layer, which has better flexibility but still high complexity. Global context network \cite{cao:gcnet} simplifies the non-local network to a lightweight structure with almost no performance loss.

\paragraph{Dynamic CNN Architectures.} There are two main stream methods for dynamic CNN architecture design. One is dynamic network structure, such as PP-LCNet \cite{cui:pplcnet}, which dynamically conducts different operations based on the inputs of different sizes in different positions of the network. And the other is dynamic convolution kernel, such as dynamic convolution \cite{chen:dyconv} and CondConv \cite{yang:condconv}, which mix different convolution kernels by generating weights through the attention mechanism. In order to exploit more efficient representations, our proposed Dite-HRNet embodies its dynamic state not only in the network structure but also in the convolution kernels.

\section{Dite-HRNet}

\begin{figure}[t]
\centering
\includegraphics[width=\linewidth]{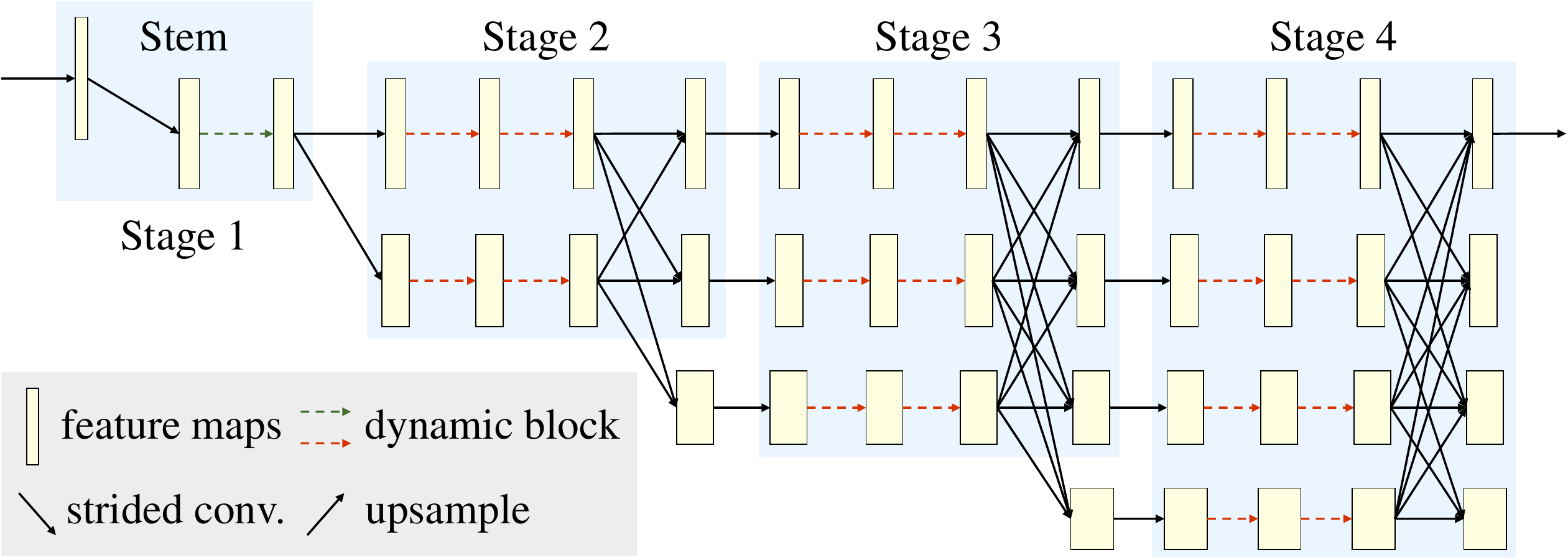}
\caption{Overall architecture of Dite-HRNet. It uses HRNet \protect \cite{sun:hrnet} as backbone, adapted by two proposed dynamic blocks. The cross-resolution modules are marked with blue areas.}
\label{fig1}
\end{figure}

\begin{table}[t]
\centering
\resizebox{\linewidth}{!}{
\begin{tabular}{llccc}
\toprule
\multirow{2}{*}[-2pt]{Stage} & \multirow{2}{*}[-2pt]{Operater type} & \multicolumn{2}{c}{\#Modules} & \multirow{2}{*}[-2pt]{\#Branches}\\
\rule{0pt}{10pt}
& & Dite-HRNet-18 & Dite-HRNet-30\\
\midrule
\multirow{2}{*}{$1^{st}$} & $3 \times 3$ strided conv ($\times 1$) & \multirow{2}{*}{1} & \multirow{2}{*}{1} & \multirow{2}{*}{1}\\
& DGC block ($\times 1$)\\
\midrule
\multirow{2}{*}{$2^{nd}$} & DMC block ($\times 2$) & \multirow{2}{*}{2} & \multirow{2}{*}{3} & \multirow{2}{*}{2}\\
& multi-scale fusion ($\times 1$)\\
\midrule
\multirow{2}{*}{$3^{rd}$} & DMC block ($\times 2$) & \multirow{2}{*}{4} & \multirow{2}{*}{8} & \multirow{2}{*}{3}\\
& multi-scale fusion ($\times 1$)\\
\midrule
\multirow{2}{*}{$4^{th}$} & DMC block ($\times 2$) & \multirow{2}{*}{2} & \multirow{2}{*}{3} & \multirow{2}{*}{4}\\
& multi-scale fusion ($\times 1$)\\
\bottomrule
\end{tabular}}
\caption{Structure of Dite-HRNet. DGC = Dynamic Global Context. DMC = Dynamic Multi-scale Context. \# = number of. $\times$ denotes repetitions of each operator in one cross-resolution module.}
\label{tab1}
\end{table}

As shown in Figure \ref{fig1}, Dite-HRNet is a 4-stage network, consisting of one high-resolution main branch with the highest resolution and three high-to-low resolution branches that are added to the network one by one in parallel at the beginning of each new stage. Each newly added branch has half the resolution and twice the number of channels compared to the previously added branch. As shown in Table \ref{tab1}, among all four stages of Dite-HRNet, the first stage, also regarded as the stem, contains a $3 \times 3$ strided convolution and a Dynamic Global Context (DGC) block on the main branch. Each subsequent stage consists of a series of cross-resolution modules which are composed of two Dynamic Multi-scale Context (DMC) blocks and a multi-scale fusion layer that exchanges information across all branches. The main branch with the highest resolution maintains a high-resolution representation, which provides the final output of the backbone network for the subsequent pose estimation. For fair comparisons, we presents two instantiations of our network, Dite-HRNet-18 and Dite-HRNet-30, which have the network widths and depths corresponding to Lite-HRNet-18 \cite{yu:litehrnet} and Lite-HRNet-30 \cite{yu:litehrnet}, respectively.


\begin{figure}[t]
\centering
\includegraphics[width=\linewidth]{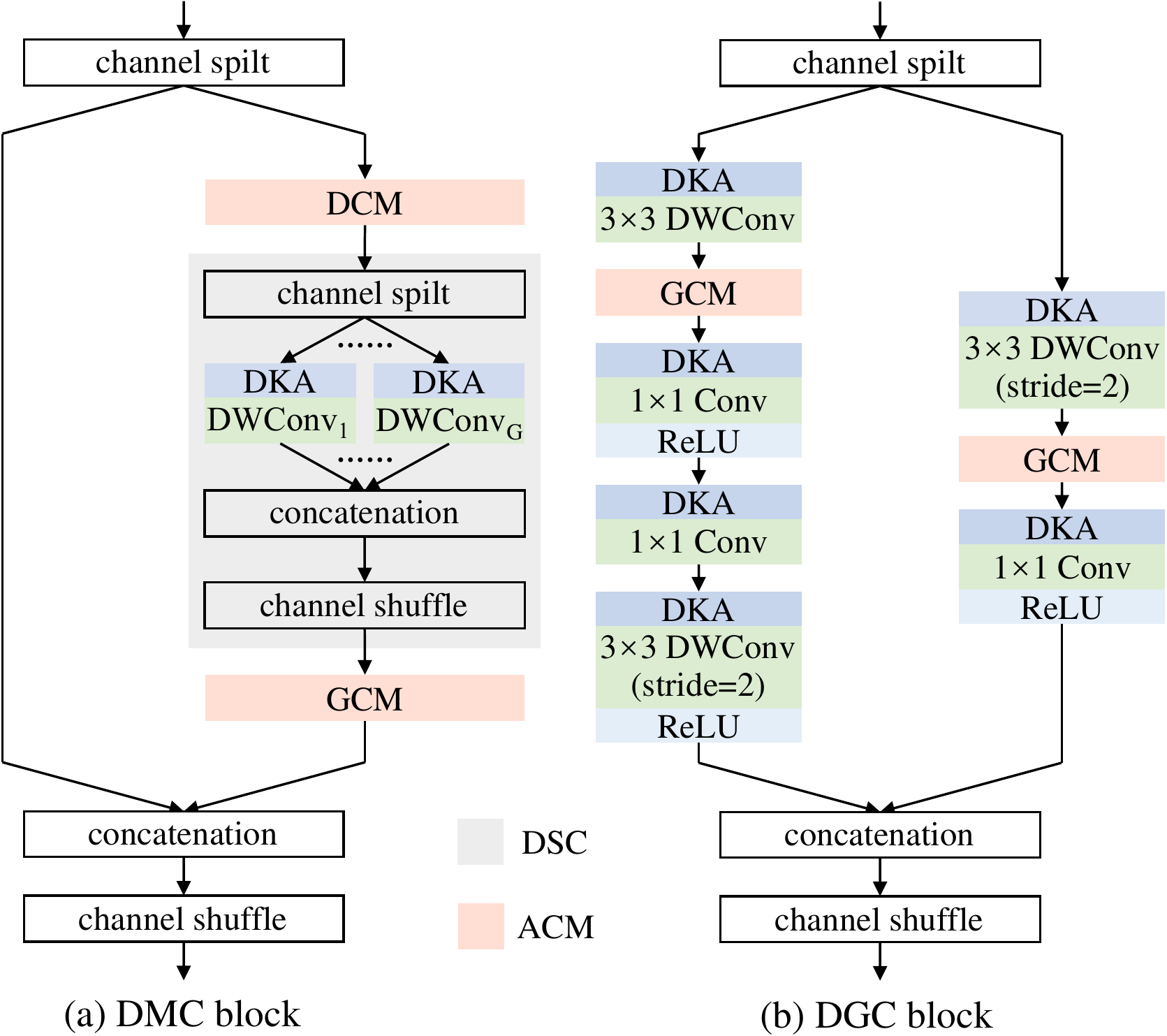}
\caption{Structures of Dynamic Multi-scale Context (DMC) block and Dynamic Global Context (DGC) block. Dynamic Kernel Aggregation (DKA) is applied to Dynamic Split Convolution (DSC) in the DMC block and each convolution in the DGC block. DCM (Dense Context Modeling) and GCM (Global Context Modeling) are two instantiations of our proposed Adaptive Context Modeling (ACM).}
\label{fig2}
\end{figure}

\paragraph{Block Design.} As shown in Figure \ref{fig2}, our DMC block and DGC block share similar overall structure, applying the channel split, feature concatenation and channel shuffle operation from ShuffleNetV2 \cite{ma:shuffle2} to aggregate different features extracted by different layers. One difference between the two blocks is that the DMC block applies a sequence of layers on half of the channels, while the DGC block applies two different sequences of layers on all two groups of channels. The sequence of layers in the DMC block contains one Dense Context Modeling (DCM) operation, one DSC and one Global Context Modeling (GCM). Both the DCM and GCM are the instantiations of the ACM method. In the DGC block, one $3 \times 3$ strided depth-wise convolution, one GCM and one $1 \times 1$ convolution are performed on one group of channels, while one $3 \times 3$ depth-wise convolution, one GCM, one $1 \times 1$ convolution and one $3 \times 3$ strided depth-wise convolution are performed on the other group of channels. Each convolution in the DGC block and the DSC layer generates the convolution kernel via Dynamic Kernel Aggregation (DKA).

\paragraph{Configuration.} In order to demonstrate the superior efficiency of our method, we optimize our networks, which are characterized by two hyper-parameters in DSC, to have similar model sizes and computational cost like in Lite-HRNet \cite{yu:litehrnet}. We configure the hyper-parameters separately on different resolution branches. Our default configuration choice is shown in Table \ref{tab5} and used for the subsequent comparison experiments. 

\subsection{Dynamic Split Convolution (DSC)}

\paragraph{Split-Concat-Shuffle (SCS) Module.} Large convolution kernels can bring broad receptive fields, thereby enhancing the long-range spatial dependency. However, using large kernel sizes on all filters in the network not only leads to a high computational cost, but also fails to capture the local pixel-wise information. Therefore, we introduce the SCS module to extract contextual information by multiple kernels of different sizes and integrate them in a single convolution layer.

In the SCS module, channels are first split into multiple groups equally, and depth-wise convolutions with different kernel sizes are applied to each group of channels in parallel. The output of the convolution on each group is formally defined as follows:
\begin{equation}
    \begin{split}
        Y_i = DWConv(K_i \times K_i|C/G) & (X_i),\\
        K_i = 2i + 1, \quad i \in [1, \ G] & ,
    \end{split}
\end{equation}%
where $X_i$ and $Y_i$ denote the input and output of the depth-wise convolution on the $i^{th}$ group of channels, respectively. $DWConv(K_i \times K_i|C/G)(\cdot)$ is the depth-wise convolution with the kernel size $K_i \times K_i$ and channel dimension $C/G$, where $C$ denotes the total number of channels among the groups and $G$ denotes the number of groups. 

After the depth-wise convolutions, the grouped features are concatenated together. To further integrate the separated information at different scales, the channel shuffle operation \cite{ma:shuffle2} is used at the bottom of the SCS module. 

Note that the SCS module does not expand the width of the network, as it only splits channels into different groups and performs different convolution operations on them in parallel. The computational efficiency is acceptable and can be optimized by the hyper-parameter $G$.

\paragraph{Dynamic Kernel Aggregation (DKA).} To make the SCS module learn rich contextual information even with small convolution kernels, we introduce a DKA operation which strengthens the input-dependency of convolution kernels by dynamically aggregating multiple kernels via kernel-wise attention weights based on the input images.

Standard convolution kernel is defined by a weight matrix $w$ with $4$ dimensions which decide the kernel size and input/output channels, respectively. Instead of concatenating the output features of different convolutions, we aggregate the kernel weight matrices $\{w_i\}$ before calculating the convolution results, to dynamically generate different convolution kernels for different inputs. The DKA operation computes attention weights over different convolution kernels, and then applies the element-wise product to the attention weights and the kernel weights. We define the DKA operation as follows:
\begin{equation}
    \begin{split}
        Y = W^T(X)X, &\\
        W(X) = \sum^{N}_{i}a_i(X & )w_i,
    \end{split}
\end{equation}%
where $a_i(X)$ is the attention weight for the $i^{th}$ convolution kernel, and $W(X)$ is the aggregated weight matrix of $N$ convolution kernels. The input-dependent attention weights $a(X)$ are computed from the input $X$ as follows:
\begin{equation}
    a(X) = Sigmoid(FC(ReLU(FC(GAP(X))))),
\end{equation}%
where $GAP(\cdot)$ represents the global average pooling, and $FC(\cdot)$ represents the full-connected layer. Two functions $Sigmoid(\cdot)$ and $ReLU(\cdot)$ are used right after two full-connected layers for non-linear activation.

Since the DKA operation takes place before the calculation of convolution results, the aggregated kernel performs only one convolution operation on each input feature map without expanding the network width. For a standard convolution with kernel size $K \times K$ and input size $C \times H \times W$, DKA of $N$ kernels only introduces minor additional computational cost $HWC + C^2/4 + CN/4$ from the standard convolution cost $HWC^2K^2$. The number of aggregated kernels $N$ is another hyper-parameter for the efficiency optimization.

\paragraph{Complementation.} The high computational efficiency makes the DKA operation an excellent complementation to the SCS module. Therefore, we combine the above two methods into one single convolution, which is named DSC. With two hyper-parameters $G$ and $N$ from the SCS module and the DKA operation, respectively, it is relatively easier for DSC to optimize the trade-off between the capacity and complexity of the convolution layers.

\subsection{Adaptive Context Modeling (ACM)}

\begin{figure}[t]
\centering
\includegraphics[width=\linewidth]{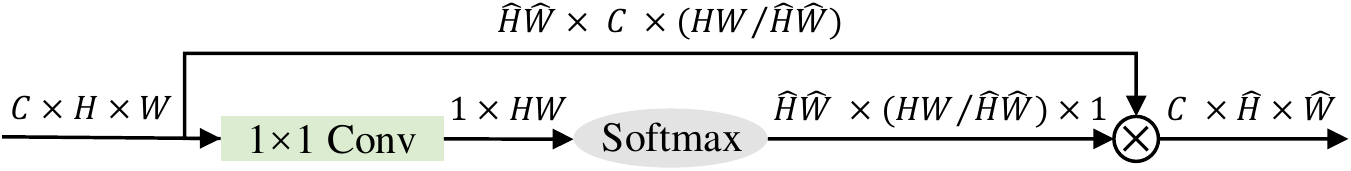}
\caption{Procedure of adaptive context pooling. $C$ denotes the number of channels. $H$ and $W$ denote the current height and width of features. $\widehat{H}$ and $\widehat{W}$ denote the target height and width of features. $\otimes$ denotes matrix multiplication.}
\label{fig3}
\end{figure}

The ACM method can be abstracted into following three steps: (a) adaptive context pooling, as shown in Figure \ref{fig3}, which creates a shortcut connection with a $1 \times 1$ channel pooling convolution and a softmax layer to form a context mask, and then applies the mask on the feature maps through a series of resolution-adaptive transformations to obtain spatial contextual features; (b) context shifting, which realigns the spatially related contextual features together by two $1 \times 1$ convolutions with non-linear activation; (c) context weighting, which adopts the element-wise weighting operation between input features and shifted context features to model the contextual relationships of the corresponding features. This abstraction of the ACM method can be defined as:
\begin{equation}
    Y = Weight(X, \ Shift(ACPool(H, \ W)(X))),
\end{equation}%
where $ACPool(H, \ W)(X)$ denotes the adaptive context pooling which pools the input features $X$ to a certain output size $H \times W$, $Shift(\cdot)$ denotes the context shifting, and $Weight(\cdot)$ denotes the context weighting.

For the high-resolution network, we propose two instantiations of the ACM method, DCM and GCM, which take the advantage of the parallel multi-resolution architecture.

\paragraph{Dense Context Modeling (DCM).} We introduce a DCM operation to densely model the spatial contextual relationships of the features from all resolution branches of one stage. At the $n^{th}$ stage, the input features from all $n$ branches are pooled to the lowest resolution $H_n \times W_n$. Then, all pooled features are concatenated together, so that the context shifting can be densely performed on the parallel contextual features. In the end, the shifted contextual features are upsampled to the corresponding resolutions and distributed back to the corresponding branches for the subsequent context weighting. This instantiation implements the ACM as:
\begin{equation}
    \begin{split}
    \resizebox{.91\linewidth}{!}{$
        \displaystyle
        \begin{cases}
            \widehat{X}_k = ACPool(H_n, \ W_n)(X_k),\\
            \widetilde{X} = Shift(Cat([\widehat{X}_1, \ ..., \ \widehat{X}_{n-1}, \ X_n])),\\
            Y_k = 
            \begin{cases}
                Weight(X_k, \ Upsamp(\widetilde{X}_k)), & 1 \leq k \leq n - 1,\\
                Weight(X_k, \ \widetilde{X}_k), & k = n,
            \end{cases}
        \end{cases}
    $}
    \end{split}
\end{equation}%
where $Cat(\cdot)$ and $Upsamp(\cdot)$ denote the feature concatenation and upsampling, respectively. $X_k$ represents the input tensor with the $k^{th}$ highest resolution. $\widehat{X}$ represents the pooled tensor from the $k^{th}$ branch. $\widetilde{X}_k$ represents the shifted tensor, which is distributed to the $k^{th}$ branch as $\widetilde{X}_k$. $Y_k$ represents the corresponding $k^{th}$ output tensor.

\paragraph{Global Context Modeling (GCM).} To separately model the global spatial dependency at each resolution, we apply a GCM operation on each branch of the network. It is a instantiation of the ACM when the adaptive context pooling has the output size $1 \times 1$. The output features of the GCM operation on the $k^{th}$ branch is defined as follows:
\begin{equation}
    \begin{split}
        Y_k = Weight(X_k, \ & Shift(ACPool(1, \ 1)(X_k))),\\
        & 1 \leq k \leq n.
    \end{split}
\end{equation}%

The GCM operation captures the spatial relationship of all features with the same resolution in the global aspect containing abundant contextual information, while the DCM operation captures the spatial relationships of all features with different resolutions in a moderate aspect containing more pixel-wise information. Meanwhile, both two operations increase the information exchange across features, so that can be better substitutes for the $1 \times 1$ convolutions in the shuffle block \cite{ma:shuffle2} than the channel weighting operations in Lite-HRNet \cite{yu:litehrnet}.

\section{Experiments}

\begin{table*}[t]
\centering
\resizebox{\linewidth}{!}{
\begin{tabular}{llclrccccccc}
\toprule
Method & Backbone & Pretrain & Input size & \#Params (M) & GFLOPs & AP & AP$^{50}$ & AP$^{75}$ & AP$^M$ & AP$^L$ & AR\\
\midrule
\multicolumn{12}{c}{Large networks}\\
\midrule
8-stage Hourglass \cite{newell:hourglass} & Hourglass & N & $256 \times 192$ & 25.1 & 14.3 & 66.9 & -- & -- & -- & -- & --\\

CPN \cite{chen:cpn} & ResNet-50 & Y & $256 \times 192$ & 27.0 & 6.2 & 68.6 & -- & -- & -- & -- & --\\

SimpleBaseline \cite{xiao:simple} & ResNet-50 & Y & $256 \times 192$ & 34.0 & 8.9 & 70.4 & 88.6 & 78.3 & 67.1 & 77.2 & 76.3\\

HRNet \cite{sun:hrnet} & HRNet-W32 & N & $256 \times 192$ & 28.5 & 7.1 & 73.4 &  89.5 & 80.7 & 70.2 & 80.1 & 78.9\\

UDP \cite{huang:udp} & HRNet-W32 & Y & $256 \times 192$ & 28.7 & 7.1 & 75.2 & 92.4 & 82.9 & 72.0 & 80.8 & 80.4\\
\midrule
\multicolumn{12}{c}{Small networks}\\
\midrule
MobileNetV2 $1 \times$ \cite{sandler:mobile2} & MobileNetV2 & N & $256 \times 192$ & 9.6 & 1.4 & 64.6 & 87.4 & 72.3 & 61.1 & 71.2 & 70.7\\

ShuffleNetV2 $1 \times$ \cite{ma:shuffle2} & ShuffleNetV2 & N & $256 \times 192$ & 7.6 & 1.2 & 59.9 & 85.4 & 66.3 & 56.6 & 66.2 & 66.4\\

Small HRNet \cite{wang:smallhrnet} & HRNet-W18 & N & $256 \times 192$ & 1.3 & 0.5 & 55.2 & 83.7 & 62.4 & 52.3 & 61.0 & 62.1\\

Lite-HRNet \cite{yu:litehrnet} & Lite-HRNet-18 & N & $256 \times 192$ & \textbf{1.1} & \textbf{0.2} & 64.8 & 86.7 & 73.0 & 62.1 & 70.5 & 71.2\\

& Lite-HRNet-30 & N & $256 \times 192$ & 1.8 & 0.3 & 67.2 & 88.0 & 75.0 & 64.3 & 73.1 & 73.3\\

\textbf{Dite-HRNet (Ours)} & \textbf{Dite-HRNet-18} & N & $256 \times 192$ & \textbf{1.1} & \textbf{0.2} &  \underline{65.9} & \underline{87.3} & \underline{74.0} & \underline{63.2} & \underline{71.6} & \underline{72.1}\\

& \textbf{Dite-HRNet-30} & N & $256 \times 192$ & 1.8 & 0.3 & 68.3 & 88.2 & 76.2 & 65.5 & 74.1 & 74.2\\
\midrule
MobileNetV2 $1 \times$ \cite{sandler:mobile2} & MobileNetV2 & N & $384 \times 288$ & 9.6 & 3.3 & 67.3 & 87.9 & 74.3 & 62.8 & 74.7 & 72.9\\

ShuffleNetV2 $1 \times$ \cite{ma:shuffle2} & ShuffleNetV2 & N & $384 \times 288$ & 7.6 & 2.8 & 63.6 & 86.5 & 70.5 & 59.5 & 70.7 & 69.7\\

Small HRNet \cite{wang:smallhrnet} & HRNet-W18 & N & $384 \times 288$ & 1.3 & 1.2 & 56.0 & 83.8 & 63.0 & 52.4 & 62.6 & 62.6\\

Lite-HRNet \cite{yu:litehrnet} & Lite-HRNet-18 & N & $384 \times 288$ & \textbf{1.1} & 0.4 & 67.6 & 87.8 & 75.0 & 64.5 & 73.7 & 73.7\\

& Lite-HRNet-30 & N & $384 \times 288$ & 1.8 & 0.7 & 70.4 & 88.7 & 77.7 & 67.5 & 76.3 & 76.2\\

\textbf{Dite-HRNet (Ours)} & \textbf{Dite-HRNet-18} & N & $384 \times 288$ & \textbf{1.1} & 0.4 & \underline{69.0} & \underline{88.0} & \underline{76.0} & \underline{65.5} & \underline{75.5} & \underline{75.0}\\

& \textbf{Dite-HRNet-30} & N & $384 \times 288$ & 1.8 & 0.7 & \textbf{71.5} & \textbf{88.9} & \textbf{78.2} & \textbf{68.2} & \textbf{77.7} & \textbf{77.2}\\
\bottomrule
\end{tabular}}
\caption{Comparisons of results on the COCO val2017 set. Pretrain = pretrain the backbone on the ImageNet classification task. \textbf{Bold} indicates the best result and \underline{underline} indicates the highest score with the lowest \#Params or FLOPs.}
\label{tab2}
\end{table*}

\begin{table*}[t]
\centering
\resizebox{\linewidth}{!}{
\begin{tabular}{llcrccccccc}
\toprule
Method & Backbone & Input size & \#Params (M) & GFLOPs & AP & AP$^{50}$ & AP$^{75}$ & AP$^M$ & AP$^L$ & AR\\
\midrule
\multicolumn{11}{c}{Large networks}\\
\midrule
SimpleBaseline \cite{xiao:simple} & ResNet-50 & $256 \times 192$ & 34.0 & 8.9 & 70.0 & 90.9 & 77.9 & 66.8 & 75.8 & 75.6\\

CPN \cite{chen:cpn} & ResNet-Inception & $384 \times 288$ & -- & -- & 72.1 &  91.4 & 80.0 & 68.7 & 77.2 & 78.5\\

HRNet \cite{sun:hrnet} & HRNet-W32 & $384 \times 288$ & 28.5 & 16.0 & 74.9 & 92.5 & 82.8 & 71.3 & 80.9 & 80.1\\

UDP \cite{huang:udp} & HRNet-W32 & $384 \times 288$ & 28.7 & 16.1 & 76.1 & 92.5 & 83.5 & 72.8 & 82.0 & 81.3\\

DARK \cite{zhang:dark} & HRNet-W48 & $384 \times 288$ & 63.6 & 32.9 & 76.2 & 92.5 & 83.6 & 72.5 & 82.4 & 81.1\\
\midrule
\multicolumn{11}{c}{Small networks}\\
\midrule
MobileNetV2 $1 \times$ \cite{sandler:mobile2} & MobileNetV2 & $384 \times 288$ & 9.8 & 3.3 & 66.8 & 90.0 & 74.0 & 62.6 & 73.3 & 72.3\\

ShuffleNetV2 $1 \times$ \cite{ma:shuffle2} & ShuffleNetV2 & $384 \times 288$ & 7.6 & 2.8 & 62.9 & 88.5 & 69.4 & 58.9 & 69.3 & 68.9\\

Small HRNet \cite{wang:smallhrnet} & HRNet-W18 & $384 \times 288$ & 1.3 & 1.2 & 55.2 & 85.8 & 61.4 & 51.7 & 61.2 & 61.5\\

Lite-HRNet \cite{yu:litehrnet} & Lite-HRNet-18 & $384 \times 288$ & \textbf{1.1} & \textbf{0.4} & 66.9 & 89.4 & 74.4 & 64.0 & 72.2 & 72.6\\

& Lite-HRNet-30 & $384 \times 288$ & 1.8 & 0.7 & 69.7 & 90.7 & 77.5 & 66.9 & 75.0 & 75.4\\

\textbf{Dite-HRNet (Ours)} & \textbf{Dite-HRNet-18} & $384 \times 288$ & \textbf{1.1} & \textbf{0.4} & \underline{68.4} & \underline{89.9} & \underline{75.8} & \underline{65.2} & \underline{73.8} & \underline{74.4}\\

& \textbf{Dite-HRNet-30} & $384 \times 288$ & 1.8 & 0.7 & \textbf{70.6} & \textbf{90.8} & \textbf{78.2} & \textbf{67.4} & \textbf{76.1} & \textbf{76.4}\\
\bottomrule
\end{tabular}}
\caption{Comparisons of results on the COCO test-dev2017 set. \#Params and FLOPs are computed for pose estimation, and those for human detection are not included. \textbf{Bold} indicates the best result and \underline{underline} indicates the highest score with the lowest \#Params or FLOPs.}
\label{tab3}
\end{table*}

\begin{table}[t]
\centering
\resizebox{\linewidth}{!}{
\begin{tabular}{lccc}
\toprule
Method & \#Params (M) & GFLOPs & PCKh\\
\midrule
MobileNetV2 $1 \times$ \cite{sandler:mobile2} & 9.6 & 1.9 & 85.4\\

MobileNetV3 $1 \times$ \cite{howard:mobile3} & 8.7 & 1.8 & 84.3\\

ShuffleNetV2 $1 \times$ \cite{ma:shuffle2} & 7.6 & 1.7 & 82.8\\

Small HRNet \cite{wang:smallhrnet} & \underline{1.3} & 0.7 & 80.2\\

Lite-HRNet-18 \cite{yu:litehrnet} & \textbf{1.1} & \textbf{0.2} & 86.1\\

Lite-HRNet-30 \cite{yu:litehrnet} & 1.8 & \underline{0.4} & \underline{87.0}\\
\midrule
\textbf{Dite-HRNet-18 (Ours)} & \textbf{1.1} & \textbf{0.2} & \underline{87.0}\\

\textbf{Dite-HRNet-30 (Ours)} & 1.8 & \underline{0.4} & \textbf{87.6}\\
\bottomrule
\end{tabular}}
\caption{Comparisons of results on the MPII val set. \textbf{Bold} indicates the best result and \underline{underline} indicates the second-best result.}
\label{tab4}
\end{table}

\subsection{Implementation Details}
\paragraph{Datasets and Evaluation Metrics.} The COCO dataset \cite{lin:coco} has images over $200K$ and $250K$ person instances, each with a label of $17$ keypoints. We train our networks on the train2017 set (contains $57K$ images and $150K$ person instances), and evaluate them on the val2017 set (contains $5K$ images) and test-dev2017 set (contains $20K$ images) by the Average Precision (AP) and Average Recall (AR) scores based on Object Keypoint Similarity (OKS). To further validate our networks, we also perform experiments on the MPII Human Pose dataset \cite{andriluka:mpii}, which contains about $25K$ images with $40K$ person instances, and evaluate the accuracy by the head-normalized Probability of Correct Keypoint (PCKh) score.

\paragraph{Training.} The presented Dite-HRNet is trained on $8$ GeForce RTX 3090 GPUs, with $32$ samples per GPU. All parameters are updated by Adam optimizer with a base learning rate $2e^{-3}$. As for the data processing, we expand all human detection boxes to a fixed aspect ratio $4 \mathbin{:} 3$, and then crop the images with the detection boxes, which are resized to $256 \times 192$ or $384 \times 288$ for the COCO dataset, and $256 \times 256$ for the MPII dataset. All images are used with data augmentations, including random rotations with factor $30$, random scales with factor $0.25$, and random flippings for both the COCO and MPII datasets. In addition, extra half body transformations are performed for the COCO dataset.

\paragraph{Testing.} The two-stage top-down paradigm, which produces the person detection boxes and then predicts the person keypoints, is adopted for our testing process. The person boxes are predicted by the person detectors provided by SimpleBaseline \cite{xiao:simple} for the COCO dataset, while the standard testing strategy for the MPII dataset uses the provided person boxes. The heatmaps are estimated via 2D Gaussian, and then averaged for the original and flipped images. The highest heatvalue locations in the heatmaps are adjusted by a quarter offset in the direction from the highest response to the second-highest response, to obtain the keypoint locations.

\subsection{Results}

\paragraph{Results on the COCO val2017 Set.} As shown in Table \ref{tab2}, we compare the results of our Dite-HRNet-18 and Dite-HRNet-30 with other state-of-the-art pose estimation methods, including both large and small networks. Our networks are trained with two different input sizes $256 \times 192$ and $384 \times 288$, achieving better trade-off between the accuracy and complexity than other methods. As compared to other small networks, our Dite-HRNet-30 achieves the highest AP score of $71.5$ with the input size $384 \times 288$. In particular, Dite-HRNet exceeds Small HRNet \cite{wang:smallhrnet} by $10$ points of AP with only $40\%$ GFLOPs. As compared to the large pose networks, our Dite-HRNet also achieves comparable or even higher accuracy with far smaller model sizes and lower computational complexity. Benefiting from the abundant contextual information, Dite-HRNet outperforms Lite-HRNet \cite{yu:litehrnet} with the equivalent parameters and GFLOPs, while having the same network depths and input sizes.

\paragraph{Results on the COCO test-dev2017 Set.} In Table \ref{tab3}, we report the results of comparisons of our networks and other state-of-the-art methods. Our Dite-HRNet achieves higher efficiency than both large and small networks. As compared to Lite-HRNet-18 \cite{yu:litehrnet}, Dite-HRNet-18 improves AP by $1.5$ points and AR by $1.8$ points with the equivalent model complexity. Our Dite-HRNet-30 achieves the highest AP of $70.6$ among the small networks, outperforming SimpleBaseline \cite{xiao:simple} while having only $5\%$ parameters and $8\%$ GFLOPs. Despite the accuracy gap with other large networks, our networks are much more competitive in terms of the model size and computational complexity.

\paragraph{Results on the MPII val Set.} The results of our networks compared with other lightweight networks are reported in Table \ref{tab4}. Our Dite-HRNet-18 improves PKCh@0.5 score by $0.9$ points over Lite-HRNet-18 \cite{yu:litehrnet} with the equivalent model complexity, has the same score but only half of GFLOPs compared to Lite-HRNet-30 \cite{yu:litehrnet}, and outperforms MobileNetV2 \cite{sandler:mobile2}, MobileNetV3 \cite{howard:mobile3}, ShuffleNetV2 \cite{ma:shuffle2} and Small HRNet \cite{wang:smallhrnet} with much lower parameters and GFLOPs. With the same model size as Lite-HRNet-30, our Dite-HRNet-30 achieves the best result of $87.6$ PKCh@0.5 among the lightweight networks.

Note that the accuracy improvements of Dite-HRNet-18 over Lite-HRNet-18 are more significant than that of Dite-HRNet-30 over Lite-HRNet-30 on all three sets. Moreover, our small version of Dite-HRNet achieves the proximate performance to the large version of Lite-HRNet. Therefore, our proposed methods are more effective for small networks, and also much more efficient than increasing the network depth.

\subsection{Ablation Study}

\begin{table}[t]
\centering
\resizebox{\linewidth}{!}{
\begin{tabular}{cc|cccc}
\toprule    
\multicolumn{6}{c}{(a) \#Params (M) / GFLOPs}\\
\rule{0pt}{10pt}
& & \multicolumn{4}{c}{$N$ (Kernels)}\\
\rule{0pt}{10pt}
& & 1 1 1 1 & 4 4 2 1 & 1 2 4 4 & 4 4 4 4\\
\midrule
\multirow{2}{*}[-4pt]{$G$} & 1 1 1 1 & 1.0 / 0.20 & 1.1 / 0.20 & 1.2 / 0.20 & 1.2 / 0.20\\

& 1 1 2 4 & 1.1 / 0.20 & \ \ 1.1 / 0.20* & 1.2 / 0.20 & 1.2 / 0.20\\

\multirow{2}{*}[4pt]{(Groups)} & 4 2 1 1 & 1.1 / 0.23 & 1.2 / 0.24 & 1.2 / 0.24 & 1.3 / 0.24\\

& 4 4 4 4 & 1.1 / 0.25 & 1.3 / 0.25 & 1.4 / 0.25 & 1.5 / 0.25\\
\toprule
\multicolumn{6}{c}{(b) AP on the COCO val2017}\\
\rule{0pt}{10pt}
& & \multicolumn{4}{c}{$N$ (Kernels)}\\
\rule{0pt}{10pt}
& & 1 1 1 1 & 4 4 2 1 & 1 2 4 4 & 4 4 4 4\\
\midrule
\multirow{2}{*}[-4pt]{$G$} & 1 1 1 1 & 65.0 & 65.4 & 65.3 & 65.3\\

& 1 1 2 4 & 65.3 & \ \ 65.9* & 65.7 & 65.8\\

\multirow{2}{*}[4pt]{(Groups)} & 4 2 1 1 & 65.5 & 65.4 & 65.3 & 65.7\\

& 4 4 4 4 & 65.8 & 66.1 & 65.8 & 65.9\\
\bottomrule
\end{tabular}}
\caption{Results of Dite-HRNet-18 with different configurations of two hyper-parameters in DSC, the number of groups $G$ and the number of kernels $N$. The numbers in the header of each row or column denote the values of $G$ or $N$ on the highest to lowest resolution branches, respectively. The results of the model with the default configuration we choose are marked with *. The GFLOPs is computed with the input size $256 \times 192$.}
\label{tab5}
\end{table}

\begin{table}[t]
\centering
\resizebox{\linewidth}{!}{
\begin{tabular}{lccccc}
\toprule
\multirow{2}{*}[-2pt]{Method} & \#Params & \multicolumn{2}{c}{COCO} & \multicolumn{2}{c}{MPII}\\
\rule{0pt}{10pt}
& (M) & MFLOPs & AP & MFLOPs & PCKh\\
\midrule
Lite-HRNet-18 \cite{yu:litehrnet} & 1.1 & 205.2 & 64.8 & 273.4 & 86.1\\

with ACM & 1.0 & 210.6 & 65.1 & 280.6 & 86.5\\

with DSC & 1.2 & 211.0 & 65.3 & 281.1 & 86.5\\

with ACM \& DSC & 1.1 & 215.5 & 65.6 & 287.2 & 86.7\\

\textbf{Dite-HRNet-18 (Ours)} & 1.1 & 209.8 & \textbf{65.9} & 279.5 & \textbf{87.0}\\
\bottomrule
\end{tabular}}
\caption{Ablation studies on the COCO val2017 and MPII val sets. The MFLOPs is computed with the input size $256 \times 192$ for the COCO val2017 set and $256 \times 256$ for the MPII val set, respectively. }
\label{tab6}
\end{table}

\paragraph{Hyper-parameters in DSC.} To explore the best configuration of two hyper-parameters $G$, $N$ in DSC, we conduct a 16-group ablation study on the COCO val2017 set. The results of our Dite-HRNet-18 with different configurations are shown in Table \ref{tab5}. On each branch of Dite-HRNet-18, parameters and GFLOPs increase as $N$, $G$ increase. However, large $N$ for the upper branches as well as large $G$ for the lower branches have less impact on the model size and computational complexity, and also achieve more improvement on accuracy. It shows a better trade-off between the accuracy and complexity by setting $G$ to $1$, $1$, $2$, $4$ and $N$ to $4$, $4$, $2$, $1$ for the highest to lowest resolution branches, respectively.

\paragraph{Effectiveness of ACM and DSC.} To further verify the effectiveness of our proposed methods, we perform ablation studies on the COCO val2017 and MPII val sets. We first use Lite-HRNet-18 \cite{yu:litehrnet} as a baseline, to which our ACM and DSC are applied separately or together, and then compare the results with our Dite-HRNet-18. Table \ref{tab6} shows that our ACM and DSC methods bring considerable accuracy improvements to Lite-HRNet-18 with negligible additional complexity for the overall model. Further, our lightweight blocks helps Dite-HRNet-18 achieve the best results and reduce the model complexity.

\section{Conclusion}
In this paper, in order to address the problems that previous high-resolution networks were input-independent and lacked long-range information, we presented a Dite-HRNet to dynamically extract feature representations for human pose estimation. Our network achieved impressive efficiency on both COCO and MPII human pose estimation datasets, due to the effectiveness of the DMC and DGC blocks, which performed input-dependent convolutions by embedding DSC and captured long-range information by embedding ACM. The proposed dynamic lightweight blocks can further expand to other networks that have multi-scale representations.

\section*{Acknowledgments}
This work was partially supported by the National Science Fund for Distinguished Young Scholars of China under Grant No.62125203, the Key Program of the National Natural Science Foundation of China under Grant No.61932013, the National Natural Science Foundation of China under Grant No. 61906099 and No. 61906098.

\bibliographystyle{named}
\bibliography{Dite-HRNet}

\end{document}